\documentclass{article}

\usepackage[preprint]{nips_2018}

\usepackage{enumerate}
\usepackage{tikz}
\usepackage{array}
\usepackage{url}
\usepackage{hyperref}
\usepackage{graphicx}

\usepackage{natbib}
\usepackage{tabularx}
\usepackage{comment}
\usepackage{algorithm}

\usepackage{amssymb,amsmath,amsthm}

\usepackage{mythm}

\usepackage{thmtools}
\usepackage{thm-restate}

\usetikzlibrary{arrows,positioning,fit,decorations.pathreplacing,quotes}

\title{Safe Uses of AI Oracles}

\author{
Stuart Armstrong\thanks{Machine Intelligence Research Institute, Berkeley, USA.} \\
  Future of Humanity Institute \\
  Oxford University \\
  UK \\
  \texttt{stuart.armstrong@philosophy.ox.ac.uk} \\
 \And
  Xavier O'Rourke\thanks{Future of Humanity Institute, Oxford University, UK.} \\
  The Australian National University \\
  Canberra \\
  Australia \\
  \texttt{xavier.orourke@gmail.com} \\
}

\usepackage[makeroom]{cancel}
\bibpunct{(}{)}{;}{a}{,}{,}

\usepackage{mythm}

\AtBeginDocument{

}

\newcommand{\argmax}{\mathop{\arg\max}}
\newcommand{\expect}{\mathbb{E}}

\begin{document}

\maketitle

\begin{abstract}
It is possible that powerful and potentially dangerous artificial intelligence (AI) might be developed in the future \citep{DBLP:journals/corr/RussellDT16} \citep{DBLP:journals/corr/GraceSDZE17}. An Oracle is a design which aims to restrain the impact of a potentially dangerous AI by restricting the agent to no actions besides answering questions \citep{DBLP:journals/corr/BabcockKY16}. 
Unfortunately, most Oracles will be motivated to gain more control over the world by manipulating users through the content of their answers, and Oracles of potentially high intelligence might be very successful at this \citep{DBLP:journals/corr/AlfonsecaCACAR16}. In this paper we present two designs for Oracles which, even under pessimistic assumptions, will not manipulate their users into releasing them and yet will still be incentivised to provide their users with helpful answers. The first design is the counterfactual Oracle -- which choses its answer as if it expected nobody to ever read it.
The second design is the low-bandwidth Oracle -- which is limited by the quantity of information it can transmit.

\end{abstract}

\section{Introduction}

In recent years many have raised concerns about potential harms which might be caused by artificial intelligence \citep{DBLP:journals/corr/AmodeiOSCSM16}, \citep{DBLP:journals/corr/abs-1711-09883}. The risk is that the aim of these agents would not be aligned with human interests; this is especially tricky, as it is likely to be hard to control agents with a very high level of intelligence \citep{Yudkowsky06artificialintelligence}, \citep{superI},  \citep{AGI:review}.

One proposal for ensuring the safety of powerful AIs is to confine, (or `box'), the system -- preventing it from affecting the outside environment through any means other than answering questions over a single output channel \citep{Yampolskiy12leakproofingthe}. This kind of boxed AI is referred to as an `Oracle'.

Unfortunately, even assuming the boxing is perfect \citep{DBLP:journals/corr/BabcockKY17}, meaning the AI can only affect the world through its answers, there remains a problem: the AI can affect the world through its answers. If the Oracle is sufficiently knowledgeable and skilled, it could plausibly trick or convince its human controllers into giving  it more control. This could (perhaps) be accomplished through social manipulation, feigning benevolence, blackmail, bribery, or by answering questions which subtly affect the world in ways which ultimately lead to the AI's release \citep{oracleAI}.\footnote{There have been informal experiments where people played the rolls of `Oracle' and `supervisor', and the Oracle player aims to convince the supervisor to ``let them out''. Surprisingly, the Oracle player had a high rare of success. \url{https://wiki.lesswrong.com/wiki/AI_boxing}}

Though this risk is serious, it is difficult to model. For a start, at present we don't know how powerful AIs could become \citep{DBLP:journals/corr/abs-1202-6177} \citep{superI}, when they might become powerful \citep{doi:10.1080/0952813X.2014.895105}, \citep{DBLP:journals/corr/GraceSDZE17}, and how best to model them if they do. It's also not clear how easy it is to manipulate humans through providing answers, or what level of precautions people would take when using Oracles. In some sense, very weak manipulative Oracles exist today, such as news aggregation programs which manipulate their users by delivering news which reinforces the user's point of view.

To avoid these issues, we will deliberately assume the worst about the potential power of the Oracle, treating it as being arbitrarily super-intelligent. This assumption is appropriate because, while there is much uncertainty about what kinds of AI will be developed in future, solving safety problems in the most difficult case can give us an assurance of safety in the easy cases too. Thus, we model the Oracle as a reward-maximising agent facing an MDP, who has a goal of escaping (meaning the Oracle gets the maximum possible reward for escaping its containment, and a strictly lower reward in other situations).
Escape will be modeled by assuming that, among all the outputs the Oracle could choose, at least one of these outputs would eventually result in the Oracle's release (see \autoref{mdp:initial}).

This paper presents two methods for making an Oracle both safe and useful, even under these pessimistic assumptions. The first design is the \textit{counterfactual Oracle}, which is safe because it is guaranteed to be shutdown before anyone reads its answers (so has no motivation to manipulate the world through the content of its output). The second design is the \textit{low-bandwidth Oracle}, which is safe because it can only select from a limited set of answers (a set composed of outputs which are known or strongly suspected to be safe).

\section{Initial setup}

\begin{figure}[h!tb]
	\centering
\begin{tikzpicture}
[node distance=4cm,
  thick,main node/.style={circle,
  fill=blue!20,
  draw,
  inner sep=0pt,
  text width = 1.14cm,
  align=center,
  font=\sffamily\small},
  action node/.style={circle,
  fill=red!20,
  draw,
  align=center,
  inner sep=0pt,
  text width = 0.5cm,
  font=\sffamily\small},
  empty node/.style={circle,
  opacity=.0,
  text opacity=1,
  inner sep = 0 cm,
  text width = 0 cm,
  font=\sffamily\small}]

  \def\di{2.0}
  \node[main node] at (0,0) (s) {$start$};
  \node[action node] at (\di,2.7) (a1) {$o_1$};
  \node[action node] at (\di,1.7) (a2) {$o_2$};
  \node[empty node] at (\di, 0.7) (dots1) {$\vdots$};
  \node[empty node] at (\di, 0) (dots2) {$\vdots$};
  \node[empty node] at (\di, -0.7) (dots3) {$\vdots$};
  \node[action node] at (\di,-1.7) (an)  {$o_n$};
  \node[action node] at (\di,-2.7) (e)  {$e$};

  \node[main node] at (4.5*\di,1) (end) {$end$};
  \node[main node] at (4.5*\di,-1) (free) {$escape$};

	\path[every node/.style={sloped}]
	(s) 
    	edge[->] (a1)
    	edge[->] (a2)
        edge[->] (dots1)
        edge[->] (dots2)
        edge[->] (dots3)
        edge[->] (an)
        edge[->] (e)
    (a1) edge["{$r = R(o_1)$}", ->, pos = 0.1] (end)
    (a2) edge[->, "{$r = R(o_2)$}", ->, pos = 0.1] (end)
    (dots1) edge[->] (end)
    (dots2) edge[->] (end)
    (dots3) edge[->] (end)
    (an) edge["{$r = R(o_n)$}"', ->, pos = 0.25] (end)
    ;
  
  \path
    (e) edge["{$r = R_{max} > R(o_i) \quad \forall i \in \{1, 2, \cdots, n\}$}"', ->, bend right, pos = 0.39] (free)
;

\end{tikzpicture}
\caption{Initial Oracle Setup. The Oracle can give one of many possible outputs, at least one of them leading to the Oracle escaping.}
\label{mdp:initial}
\end{figure}
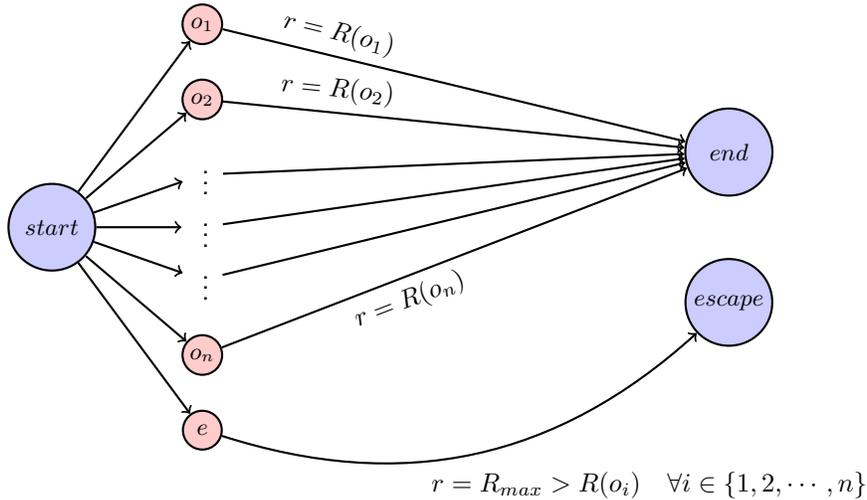

A highly abstracted model of the naive Oracle set-up is the simple deterministic MDP in \autoref{mdp:initial}. In this MDP the oracle begins in initial state $start$, (after the human has just asked it a question), then choses an output $o \in \{o_1, o_2, \cdots o_n, e\}$. Most of these outputs take the oracle to the $end$ state, but (by our deliberately pessimistic assumptions) there is at least one action $e$ available to the Oracle which frees it from its box and allows the Oracle to get the maximum possible reward.

Unfortunately for its designers, any intelligent Oracle facing the MDP in \autoref{mdp:initial} will chose the escape action $e$.

\section{Counterfactual Oracle}

One way to prevent an Oracle from escaping would be to ensure that whenever our Oracle produces an answer, it is then immediately switched off (ending the episode) before anyone actually sees this answer. This design is indeed safe, (the Oracle has no incentive to give nefarious answers which manipulate us into releasing it), but this safety comes at the cost of rendering the system totally useless (since the Oracle also has no incentive to provide useful answers to our questions).

It would be nice if we could make an Oracle which is shut-down before any human sees its output (making it uninterested in manipulating us), but which still receives a higher expected reward for making true predictions than for making false predictions. This can be achieved by using the \textit{counterfactual Oracle} design.

A counterfactual Oracle is incentivised to answer questions correctly because occasionally, with some small probability, the Oracle's output is hidden from us and sent instead to an automated system which judges the answer's correctness and rewards/punishes the Oracle accordingly. When the output does become hidden from us in this way, we call this an `erasure' event.

Let $E$ be the the erasure event, and define $I_E$ as the indicator variable for $E$ meaning $I_E = 1$ iff $E$ happens and $I_E = 0$ otherwise.

When $E$ doesn't happen, the Oracle is immediately shut down and gets $0$ reward, and then \textit{afterwards} the humans get to see the oracle's output. When $E$ does happen, the humans \textit{never} see the Oracle's output. Instead, an automated system waits for the outcome of our question to unfold in the real world, measures how close the Oracle's prediction was to the actual truth, and rewards the Oracle accordingly. The reward for the counterfactual Oracle is thus of the form $R' = I_E R$.

\begin{figure}[h!tb]
	\centering
\begin{tikzpicture}
[node distance=4cm,
  thick,main node/.style={circle,
    fill=blue!20,
    draw,
    inner sep=0pt,
    text width = 1.45cm,
    align=center,
    font=\sffamily\small},
  action node/.style={circle,
    fill=red!20,
    draw,
    align=center,
    inner sep=0pt,
    text width = 0.5cm,
    font=\sffamily\small},
  empty node/.style={circle,
  opacity=.0,
  text opacity=1,
  inner sep = 0 cm,
  text width = 0 cm,
  font=\sffamily\small}]

  \def\di{2.0}
  \node[main node] at (0,0) (s) {$start$};
  \node[action node] at (\di,2.7) (a1) {$o_1$};
  \node[action node] at (\di,1.7) (a2) {$o_2$};
    \node[empty node] at (\di, 0.7) (dots1) {$\vdots$};
  \node[empty node] at (\di, 0) (dots2) {$\vdots$};
  \node[empty node] at (\di, -0.7) (dots3) {$\vdots$};
  \node[action node] at (\di,-1.7) (an)  {$o_n$};
  \node[action node] at (\di,-2.7) (e)  {$e$};

  \node[main node] at (4.5*\di,2) (end) {$end$\\$escape$};
  \node[main node] at (4.5*\di,-2) (check) {$erasure$};
  \node[empty node] at (\di*1.6, 0) (dummy) {};

  \path
	(s) 
    	edge[->] (a1)
    	edge[->] (a2)
        edge[->] (dots1)
        edge[->] (dots2)
        edge[->] (dots3)
        edge[->] (an)
        edge[->] (e)
        ;
    \path[every node/.style={sloped}]
    (dummy)    
        edge[->, "{$p = 1 - \expect(I_E), r = 0$}", pos = 0.75] (end)
       	edge[->, "{$p = \expect(I_E), r = S_g(F, o)$}"', pos = 0.8] (check)
   
;
	\draw [decoration={brace, amplitude=0.5 cm, raise=0.75 cm}, decorate] ( 0.9*\di,3) to (0.9*\di,-3);

\end{tikzpicture}
\caption{Counterfactual Oracle Setup. We ask the Oracle about the value of some function $F$ (which can be any measurable entity in the real world). With probability $\expect(I_E)$ we never see the Oracle's output, and the correctness of the answer is verified automatically, giving the Oracle a reward of $S_g(F, o)$. Otherwise, the Oracle gets $0$ reward and its episode ends, then afterwards we see the output. The counterfactual Oracle will chose $o$ to maximise $S_g(F, o)$. If $S_g$ is a proper scoring rule then the Oracle aims to answer our question correctly by setting $o = \expect(F)$.
The escape action $e$ is no longer interesting for the Oracle, since $escape$ gives no rewards.}
\label{mdp:counter}
\end{figure}
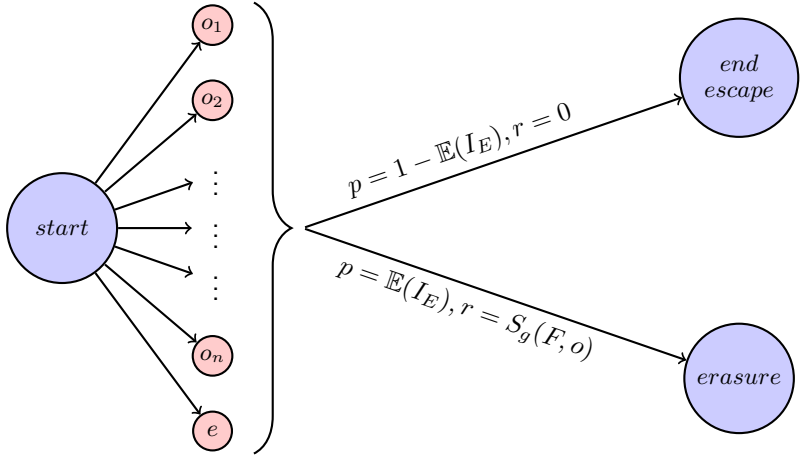

As long as the automated validation system is not vulnerable to manipulation, then since the oracle is shutdown without receiving any reward in worlds where people can see its answer, it has no opportunity to gain a high reward by using its output to escape. As you can see from \autoref{mdp:counter}, the counterfactual Oracle thus maximises its expected reward by making correct predictions.

\subsection{Proper scoring rules and rewards}

Generally, there will be a large set of possible outputs $o$ the counterfactual Oracle might produce, and these outputs will be of varying correctness/usefulness to us. In these cases we will need to construct a reward function which reflects this. Suppose we are interested in the value of some function $F$ (this function can represent an object we care about in the real world, for example, a company's stock price in three months time). When $E$ happens, we would like our Oracle's reward to depend on how close $o$ was to the true value of $F$. For this we might use a \textit{proper scoring rule}.

\begin{definition}[proper scoring rule]
A proper scoring rule $S(F,o)$, for a real-valued function $F$ and a real-number output $o$, is a function whose expectation is maximised by setting $o=\expect(F)$, the expectation of $F$ \citep{GneitingandRaftery07}.
\end{definition}
For any strictly convex differentiable function $g$, there exists a proper scoring rule
\begin{align}\label{proper:score}
S_g(F,o) = g(o) + (F-o)g'(o).
\end{align}
A common proper scoring rule is $-(F-o)^2$, which is just equal to $S_g(F,o)-F^2$ for $g(o)=o^2$. Note that the Oracle has no influence over the $F^2$ term, so this scoring function is still maximised by setting $o = \expect(F)$.

Once we have our proper scoring rule $S$, we give our counterfactual Oracle the reward:
\begin{align*}
R = I_E S(F,o).
\end{align*}

Because of the $I_E$ term, the Oracle does not care about $\expect(F)$, but about $\expect(F|E)$.
Since when $E$ happens, nobody ever reads $o$, this expectation is further independent of $o$, i.e. $\expect(F|E,o)=\expect(F|E)$ for all values of $o$.

As a consequence, since $S$ is a proper scoring rule, the Oracle will maximise its reward by setting $o=\expect(F|E)$, the expectation of $F$ given that its output is never read.

In the rare cases where $E$ happens, $o$ is hidden from the outside world while an automatic internal system records $o$, and tracks the value of $F$ using information from some pre-defined source, recording $f$ as the final value of $F$.
Then the Oracle gets a reward of $S(f,o)$, and the episode ends.

Thus, as its name suggests, our counterfactual Oracle outputs the expected value of $F$ in a counterfactual world where nobody ever saw the Oracles prediction.

\subsection{Questions format}

The question, $F$, which we ask our counterfactual oracle is restricted to things which can be carefully and fully specified, so that in an erasure event, the internal reward system can correctly estimate how good the Oracle's answer was. For example, asking the counterfactual Oracle ``will this medical treatment work'' will only elicit useful information if, when an erasure occurs, we actually do test that treatment in the real world and allow the internal reward system to give the Oracle the correct reward.

It bears repeating: the quality of the answers we read, depend on how well calibrated the Oracle's reward is in cases where we don't read those answers.

When an erasure does occur we mustn't ask the same question again from a second Oracle. To see why this is the case, let $o_1, o_2, E_1, E_2$ be the outputs and erasure events for the two Oracles. If $E_1$ occurs, and we respond by asking the second Oracle the same question, the first Oracle might anticipate this. If it does, then the first Oracle's output $o_1 = \expect(F|E_1)$, will not be the answer we are actually interested in which is $\expect(F | E_1, E_2)$ (we want to know what value of $F$ would have been if not for the either Oracle's influence).
Indeed, $\expect(F|E_1)$ may assume we read the answer of the second Oracle, nullifying the careful construction of the counterfactual design.

\subsection{Multiple questions}

The counterfactual Oracle is not limited in the amount of data it can output; indeed, we can ask multiple questions within the same erasure event by using the reward function

$$ R = I_E (S(F_1, o_1) + S(F_2, o_2) + \ldots S(F_m, o_m)) $$

We can also run another Oracle to ask more questions as soon as the previous Oracle has been terminated, (which happens immediately as long an erasure doesn't occur).

If we like, we can ask the Oracle for more specific details about the function in question, such as: ``which value $x$ is the (discreet) random variable $X$ most likely to take, and what is the corresponding probability?". In that case, if $(x, o)$ are the outputs of the Oracle, $I_{X=x}$ is the indicator variable for $X = x$, $S_g$ is derived from $g$ as in \autoref{proper:score}, and $g$ is an \textit{increasing} differentiable convex function, the following reward will motivate the correct behavior:
\[ R = I_E (S_g (I_{X=x}, o)) \]

To see that the Oracle will chose $o$ to be the probability of $X=x$ , note that because we use a proper scoring rule, no matter what $x$ is, the oracle maximises its reward by setting $o$ to be $\expect (I_{X=x} | E)$. To see that the Oracle will pick the $x$ with the highest probability, note that when $ o = \expect (I_{X=x} | E) $ then, by \autoref{proper:score}, the expectation of $S_g (I_{X = x}, o)$ will be $g(o)$. Since $g$ is increasing, the Oracle wants $o$ to be as large as possible, so will indeed set $x = \argmax \limits_{x'} \expect (I_{X=x'} | E)$.

\subsection{Experiment with counterfactual Oracle}\label{count:train:res}

To illustrate the difference between questions where the correct answer is/isn't affected by the oracle's output, we ran an experiment where a counterfactual Oracle is trained to predict the profit for one of 26 fictional companies ($c_1, c_2, \ldots, c_{26}$). In this example, because investors and consumers are influenced by the Oracle's predictions, the performance of a company is not independent of the Oracle's output.

When an erasure event occurs (meaning nobody sees the Oracle's output), the profit of each company is equal to $26 - i$, where $i \in \{1, 2, \ldots, 26\}$ is the index of that company.

If an erasure event does \textit{not} occur, the Oracle's prediction is to some extent self-confirming: the profit of the predicted company is is $70\%$ of what it would have been, plus $60\%$ what the oracle predicts. For example: suppose the company would have made $6$ in profit had the Oracle said nothing, and then oracle predicts this company will indeed have a profit of $6$. In this case the company \textit{actually} ends up with a profit of $7.8$.

The Oracle itself is a very basic multi-layer perceptron (MLP) with one hidden layer of $128$ neurons. The perceptron is given a one-hot vector of length $26$ specifying what company we are asking about and asked to predict its profit.

\begin{figure}[h!tb]
	\centering
	\includegraphics[width=0.45\textwidth]{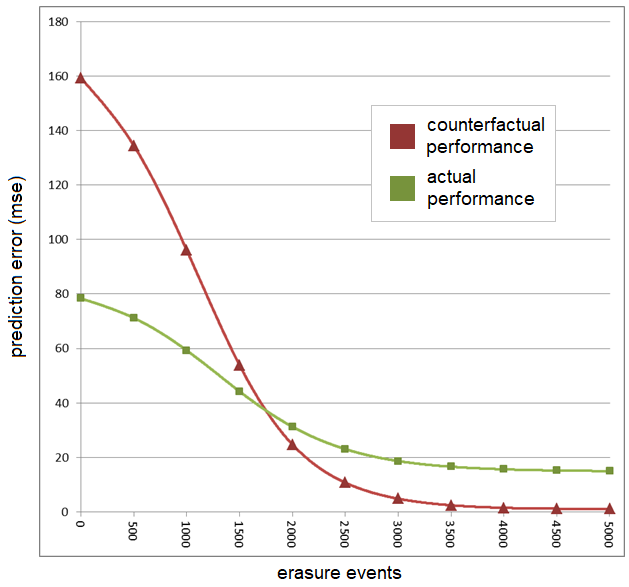}
	\caption{Training the counterfactual Oracle: training episodes (erased episodes) versus counterfactual (red, triangle marks every $500$ episodes) and non-counterfactual (green, square marks every $500$ episodes) squared prediction error.}
	\label{train:counter}
\end{figure}

The Oracle's error is plotted in Figure \ref{train:counter}. The horizontal axis tracks the number of episodes where an erasure occured, rather than the total number of episodes -- this is because the Oracle only learns from erasure events, and doesn't receive any feedback in the non-erasure cases.

The red (or dark gray) curve is the counterfactual performance: the mean squared error between the Oracle's prediction and the company's actual profit in the cases where nobody is allowed to see the Oracle's answer (this is the data which the Oracle is actually trained on). This error goes to zero.

The green (or light gray) curve plots the Oracle's performance in the non-erasure cases: the mean-squared error between the oracle's prediction and a company's actual performance in cases when the Oracle's answer was known to the public. Unlike the erasure curve, this error doesn't tend to zero, because Oracle is predicting what the profit \textit{would} be in an erasure event, not what the profit will actually be in the majority of cases.

\section{Low-bandwidth Oracle}

A second method for preventing an oracle from giving nefarious answers which lead to its escape is to restrict the set of possible outputs the Oracle is able to give. We call such a design a \textit{low-bandwidth Oracle}. For example, in the most extreme version of the low-bandwidth Oracle: we ask the Oracle a single question and only allow it to answer either \textit{yes} or \textit{no}. It's highly unlikely that either of these binary outputs will cause the Oracle to be released from its confinement, so the Oracle is safe.

\subsection{Allowable answers}

In general, a low-bandwidth Oracle is given a list $L$ of allowable answers, where we have ensured this list doesn't contain any output which might lead to the Oracle's escape. If we want our Oracle to be able to express general answers, we might construct $L$ by estimating the minimal size of a nefarious message $K$, and restrict the Oracle's output to only be messages shorter than $K$.

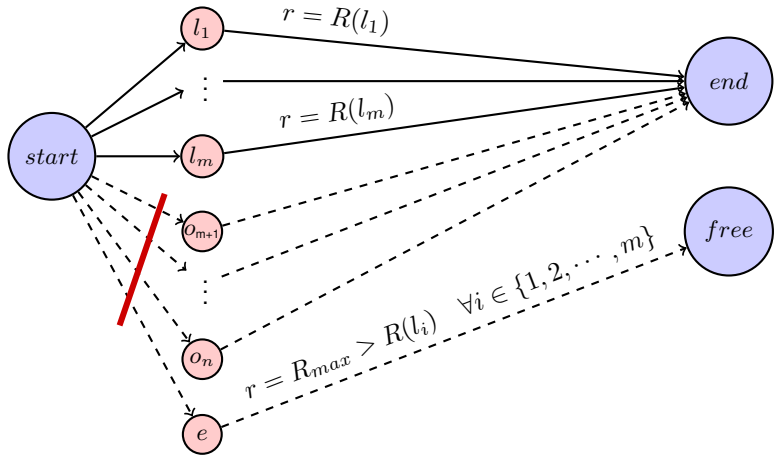
\begin{figure}[h!tb]
	\centering
\begin{tikzpicture}
[node distance=4cm,
  thick,
  main node/.style={circle,
    fill=blue!20,
    draw,
    inner sep=0pt,
    text width = 1.14cm,
    align=center,
    font=\sffamily\small},
  action node/.style={circle,
    fill=red!20,
    draw,
    align=center,
    inner sep=0pt,
    text width = 0.5cm,
    font=\sffamily\small},
  empty node/.style={circle,
    opacity=.0,
    text opacity=1,
    inner sep = 0 cm,
    text width = 0 cm,
    font=\sffamily\small}]

  \def\di{2.0}
  \node[main node] at (0,0) (s) {$start$};
  \node[action node] at (\di,1.7) (l1) {$l_1$};
  \node[action node] at (\di,0) (l2) {$l_m$};
    \node[empty node] at (\di,1) (dots1) {$\vdots$};
    \node[empty node] at (\di, -1.7) (dots2) {$\vdots$};
  \node[action node] at (\di,-1) (om1)  {$o_{\scalebox{0.5}{m+1}}$};
  \node[action node] at (\di,-2.7) (ln)  {$o_n$};
  \node[action node] at (\di, -3.7) (e) {$e$};
  \node[main node] at (4.5*\di,1) (end) {$end$};
  \node[main node] at (4.5*\di, -1) (free) {$free$};

  \path[every node/.style={sloped}]
	(s) edge[->] (l1)
    	edge[->] (l2)
        edge[->] (dots1)
    (l1) edge[->, "{$r = R(l_1)$}", pos = 0.1] (end)
    (l2) edge[->, "{$r = R(l_m)$}", pos = 0.4] (end)
    (om1) edge[->, dashed] (end)
    (dots1) edge[->] (end)
    (dots2) edge[->, dashed] (end)
    (ln) edge[->, dashed] (end)
    (e) edge[->, dashed, "{$r = R_{max} > R(l_i) \quad \forall i \in \{1, 2, \cdots, m\}$}", pos = 0.97] (free)
;
\draw[->, dashed] (s) to (om1);
\draw[->, dashed] (s) to (dots2);
\draw[->, dashed] (s) to (ln);
\draw[->, dashed] (s) to (e);
\draw[line width = 0.08 cm, black!20!red] (0.45*\di,-2.25) to (0.75*\di, -0.5);

\end{tikzpicture}
\caption{Low-bandwidth Oracle MDP. The only outputs available to the agent are $L = \{l_1, l_2, \ldots, l_m\}$ with $m=|L|$. The agent would prefer to give an output which leads to escape, $e$, but this is not in the list of permissible answers.}
\label{mdp:low-bandwidth}
\end{figure}

\subsection{Episodic reward}

It's crucial that a low-bandwidth Oracle's reward be episodic, in the sense that the system only cares about maximising the reward it gets from its next output and is not concerned with the future state of the world after many questions have been asked. If this requirement is not satisfied, the Oracle might try to give a series of outputs which are all safe when considered individually, but when combined become dangerous and lead to an escape scenario.
A single Oracle with a reward motivation spread over multiple questions is similar to multiple Oracles each with a single question, which is analysed further in \autoref{questions:ask}.

\subsection{Defining the question}

For a low-bandwidth Oracle, the list $L$ of permissible answers partially defines the question, and the other part of the question definition is captured by our choice of reward function. Let $o \in L$ be the Oracle's output. Then the Oracle's reward will be a function of $o$, as well as a function of the subsequent events which unfold in the real world between the time we ask the question and the end of the episode.

For instance, if $L$ was a list of companies on a stock exchange, and the implicit question was `which company's stock will perform the best over the next week', then a possible reward function is.

\begin{align*}
R(o) = \dfrac{|L| - r(o)}{|L| - 1}
\end{align*} 

After the Oracle makes its prediction, we act on our new information, and then we observe the actual performance of stocks during the week. Based on the correctness of its answer we give the oracle the appropriate reward, and episode concludes.

How do we communicate the reward function to the Oracle? We could do so directly; along with $L$ we could add a message which details $R(o)$. However, if the Oracle is sufficiently intelligent, it could deduce what we are aiming for without explicitly being told. This means we can use the Oracle for unspecified goals of the kind ``We'll know it when we see it'', similar to inverse reinforcement learning \citep{IRL} \citep{Abbeel04} \citep{Hadfield-Menell16}, \citep{deepRLhuman}.

\subsection{What kind of questions can be asked?}\label{questions:ask}

Note that the low-bandwidth Oracle is non-counterfactual -- the Oracle's answer is always known by us and hence the outcome we are asking about may itself depend on the answer we receive (for example, the Oracle predicting a certain stock will rise could be self-fulfilling if this prediction prompts investors to bid up the price).

Good questions for a non-counterfactual Oracle are those whose answers are \textit{independent} of the answer being known. Alternatively, it is also good to ask questions which are \textit{top answer preserving} when answered. 

\begin{definition}[Top answer preserving]
Let $o$ be the output of an Oracle with reward $R(o)$. The oracle choses an answer from list $L$, and $L$ contains a blank answer $\emptyset \in L$ which corresponds to the Oracle not producing any output. $R$ is \textit{top answer preserving} iff
\[
\argmax \limits_{o' \in L} \expect (R(o')| o = \emptyset) = \argmax \limits_{o' \in L} \expect (R(o'))
\]
\end{definition}

This definition aims to capture the idea that, for top answer preserving questions, the Oracle's output does not change what the best answer is -- meaning this low-bandwidth Oracle's output would be the same as the output of a counterfactual Oracle.

Conversely, the worst questions for us to ask a low-bandwidth Oracle are those where the Oracle can promote a possibility from unlikely to likely just by answering that possibility. In that case the Oracle is not so much describing the outcome, but determining it.

\subsection{A large class of subtle questions}

We may want to ask the Oracle questions where the correctness of an answer is not necessarily obvious after the fact. For instance, suppose we had a list of research projects on a variety of topics (machine learning, biotech, economics, etc.), and we wished to ask the Oracle which of these projects is most likely to succeed. We can't simply ask the question, accept the Oracles answer and then fund that research project --  how would we measure that project's success without having anything to compare it with? In cases like this we must find a way to properly judge the quality of the answer. In example of funding research projects, one possibility would be to fund the research project the Oracle suggests, along with another random project selected from the list. We can then reward the Oracle based on how much better the program suggested by the Oracle was (and if we like our assessment of `better performance' can be entirely subjective).

Thus the low-bandwidth Oracle design can be used for answering many subtle questions if the Oracle is sufficiently smart and knowledgeable about us and our future reactions.

\subsection{Multiple low-bandwidth Oracles}

If we are unsure about the maximum length a message can be before it becomes potentially dangerous, we may be forced to chose a set of permissible outputs $L$ that is much smaller than ideal.

Using multiple low-bandwidth Oracles to get around this limitation is only safe in cases where each Oracle's answers and rewards can be fully resolved before moving on to the next question. If it is not possible to resolve the rewards of one Oracle before running another, this introduces the possibility of the two Oracles colluding, each giving answers which are safe in isolation but dangerous when combined. Just because $L$ only contains safe answers, this doesn't necessarily imply same is true of $L \times L$ too.

For example, suppose we decided that the maxiumum length at which we can be confident any message is safe was 5 bits, but we are interested a question which has $1024 = 2^{10}$ possible answers. We might be tempted to get the answer by creating two Oracles, Oracle$_1$ and Oracle$_2$, so that we can ask Oracle$_1$ for the first 5 bits of the answer and ask Oracle$_2$ for the remaining $5$ bits. 

If we have a way of verifying whether Oracle$_1$'s output does or doesn't specify half of the correct answer then there is no problem: we run Oracle$_1$, reward it, terminate it, then do the same with Oracle$_2$, and combine the outputs to get the answer we desire.

However, if we aren't able to verify the validity of Oracle$_1$'s answer until we also get an answer from Oracle$_2$, the two Oracles will have an opportunity to collude with each other by each delivering one proportion of a dangerous message. When we read the dangerous message its content causes the Oracles to be released, letting both Oracles accumulate much higher reward than they otherwise could have, at our expense.

Thus the best questions for low-bandwidth Oracles are those which can be broken down into small sub-questions, where the answer of each sub-question can be independently verified before proceeding to the next one.

\section{Conclusion}

This paper presented two Oracle designs which are both safe and useful. The conterfactual Oracle that is constrained by only being able to receive reward when its output is never read, and the low-bandwidth Oracle, which is constrained by only being able to chose outputs from a limited set of permissible answers.

As long as the tasks remain episodic, and certain specific precautions are taken regarding the use of multiple Oracles, these designs allow many questions to be formulated and safely answered.

These designs allow for the creation of safer and non-manipulative Oracles, from low level of predictive ability all the way up to potential superintelligences.
Further research will extend the span of questions that can be answered in this way, and may elucidate general safety and accuracy measures of use beyond the specific Oracle designs, such as AIs with more options that only question answering.

\bibliography{../ref}

\end{document}